\documentclass[times, review, 10pt]{elsarticle}

\usepackage{cite}
\usepackage{amsmath,amssymb,amsfonts}
\usepackage[graphicx]{realboxes}
\usepackage{textcomp}
\usepackage[table,xcdraw]{xcolor}
\usepackage{multirow}
\usepackage{array}
\usepackage{tikz}
\usepackage{siunitx}
\usepackage{lscape}
\usepackage{adjustbox}
\usepackage{tabularx}
\usepackage{mleftright}
\usepackage{etoolbox}
\usepackage{xcolor}
\usepackage{amsmath}
\usepackage{amssymb}
\usepackage[normalem]{ulem}
\usepackage{tablefootnote}
\usepackage{verbatim}
\usepackage{graphicx}
\usepackage{booktabs}
\usepackage{mathtools}
\usepackage{makecell}
\usepackage{colortbl}
\usepackage{float}
\usepackage{listings}
\usepackage{algorithm}
\usepackage{algpseudocode}
\usepackage[dvipsnames]{xcolor}

\algtext*{EndFor}  
\algtext*{EndIf}  
\algtext*{EndWhile} 
\algtext*{EndFunction} 
\journal{Pattern Recognition}

\begin{document}

\begin{frontmatter}

\title{Multi-stage Prompt Refinement for \\Mitigating Hallucinations in Large Language Models} 

\author[label1]{Jung-Woo Shim}
\ead{j_w_shim@korea.ac.kr}
\author[label1]{Yeong-Joon Ju}
\ead{yj_ju@korea.ac.kr}
\author[label1]{Ji-Hoon Park}
\ead{jhoon_park@korea.ac.kr}
\author[label1]{Seong-Whan Lee\corref{cor1}}
\ead{sw.lee@korea.ac.kr}
\affiliation[label1]{organization={Department of Artificial Intelligence, Korea University},
            addressline={Anam-dong}, 
            state={\\ Seongbuk-gu},
            city={Seoul},
            postcode={02841}, 
            country={Republic of Korea}}
\cortext[cor1]{Corresponding Author}

\begin{abstract}

Recent advancements in large language models (LLMs) have shown strong performance in natural language understanding and generation tasks. However, LLMs continue to encounter challenges with hallucinations, where models generate plausible but incorrect information. While several factors contribute to hallucinations, the impact of ill-formed prompts, prompts with ambiguous wording, incorrect grammar, or incomplete information, was relatively under explored. To address this, we introduce Multi-stage Prompt Refinement (MPR), a framework designed to systematically improve these ill-formed prompts across multiple stages. Each stage addresses specific errors such as punctuation, typographical mistakes, and misuse of key terms, using small language models (SLMs) fine-tuned for these tasks. MPR iteratively enhances the clarity of prompts with additional context and employs a self-reflection mechanism with ranking to prioritize the most relevant input. Experimental results on hallucination benchmarks show that prompts refined by MPR achieve over an 85~\% win rate compared to their original forms, demonstrating its effectiveness in reducing hallucinations and improving LLM output accuracy. Interestingly, we reveal that MPR can be combined with existing post-hoc hallucination mitigation frameworks, further enhancing its versatility. MPR provides a lightweight and adaptable solution for enhancing LLM reliability across various domains.

\end{abstract}


\end{frontmatter}

\section{Introduction} \label{sec1}

Advancements in artificial intelligence have significantly improved large language models (LLMs), enabling them to excel in tasks like contextual understanding and human-like text generation~\citep{pr_8, pr_9, gpt4_2023, palm2_2023, pr_5}. Despite these successes, a major challenge remains: LLMs often produce hallucinations—plausible but incorrect information~\citep{llmsurvey_2023}. These hallucinations pose risks in critical fields like healthcare and education, where conveying accurate information is essential.
Current approaches to mitigate hallucinations typically focus on adjusting LLM architectures or post-processing outputs~\citep{llmlies_2023, pr_3}, often overlooking the quality of the user’s prompt. Poorly constructed prompts directly contribute to inaccurate outputs, yet existing solutions frequently rely on large models or computationally expensive techniques~\citep{optimizers_2024, factcheckgpt_2023}. These approaches also require model-specific adjustments, limiting their scalability and practicality, especially in resource-constrained environments.

To address these challenges, we introduce \textbf{Multi-Stage Prompt Refinement (MPR)}, a framework that systematically refines user prompts before they are processed by LLMs. MPR employs a three-stage process to correct ordinary punctuation, typographical, and grammatical errors along with resolving misused or ambiguous key terms.
Each stage uses specialized small language models (SLMs) fine-tuned with parameter efficient fine-tuning instead of full fine-tuning to adapt to specific tasks for correcting such errors. After refinement, MPR generates descriptions and iteratively improves them to provide comprehensive context for the LLM via self-reflection along with a ranking procedure, ensuring clearer and more accurate outputs.

Our evaluations show that MPR significantly reduces hallucinations and improves LLM response accuracy by systematically refining prompts and enriching them with detailed task-specific descriptions. This approach addresses common prompt deficiencies such as punctuation errors, typographical mistakes, and ambiguous terminology, ensuring that LLMs receive well-structured, contextually complete inputs. By enhancing the quality and precision of user prompts, MPR improves both the coherence and relevance of LLM-generated outputs, leading to more accurate and reliable responses.
One of the primary strengths of MPR is its lightweight and model-agnostic design, which allows for seamless integration with a variety of LLM architectures. This flexibility ensures that MPR can be easily adapted for different use cases, including environments with limited computational resources, where large-scale models may not be feasible. Additionally, MPR's modularity enables it to scale efficiently across various domains, making it applicable to a wide range of industries, from general-purpose applications to domain-specific fields like healthcare, legal services, and finance. Through this, MPR represents a critical advancement in improving the overall performance and reliability of large language models across diverse real-world applications.

\section{Related Works} \label{sec2}

\subsection{Challenges in Hallucination Mitigation} \label{subsecChallenges}

Mitigating hallucinations in LLMs remains challenging despite ongoing efforts. Current methods focus on modifying model architectures, fine-tuning with larger datasets, or using post-processing techniques. Adjusting internal mechanisms and using external verification tools like knowledge bases have shown promise~\citep{ hallucinationorigin_2023,pr_2}, but these approaches are computationally expensive and hard to scale across domains~\citep{preventinghallucinations_2024}.
Fine-tuning with reinforcement learning has also been effective, rewarding factual responses and penalizing errors~\citep{finetuning_2023}. However, it requires significant computational resources and is difficult to apply in real-time or resource-constrained environments due to its model-specific configurations.
Most approaches address hallucinations post-hoc; after generation, verifying content after generation. While effective, these methods add complexity, introduce latency, and assume well-structured prompts, limiting their suitability for real-time use.

\subsection{SLMs and Their Advantages} \label{subsecLLM_SLM}

In contrast to large-scale LLMs, SLMs provide a more resource-efficient option for specific tasks~\citep{distilling_2023} due to its emergent abilities obtained through knowledge distillation~\citep{slmresource_2024}. Although they lack the extensive data and capacity of LLMs, SLMs excel when fine-tuned for narrow, task-specific applications~\citep{differently_2023, pr_6}. Their computational efficiency makes them valuable in resource-constrained scenarios or when full LLM capabilities aren’t needed.
SLMs are particularly suited for real-time applications or environments with limited computing resources~\citep{knowledgeintensive_2024,pr_1}. They perform well on smaller datasets, making them ideal for tasks like prompt refinement, where the focus is on improving input quality before processing by larger models~\citep{teachingSLMs_2022,pr_7}. Research shows that SLMs fine-tuned for tasks like error correction or semantic analysis can boost LLM performance while avoiding the high computational costs of LLMs~\citep{SLMsaregood_2024}. Their efficiency and adaptability make SLMs ideal for targeted applications that enhance LLM output accuracy.

\subsection{Prompt Refinement} \label{subsecPromptRefinement}

Prompt refinement has become essential for improving input quality and reducing hallucinations in LLM outputs. Traditional strategies, such as using LLMs to rephrase or enhance prompts~\citep{promptrefinement_2023}, often compromise computational efficiency. Techniques like reinforcement learning~\citep{promptrewritingrl_2024,pr_3} and human intervention~\citep{humanfeedback_2024,pr_4} have been used to iteratively improve prompt clarity, ensuring more accurate inputs for LLMs.
Automated methods, including paraphrasing models and supplementary task descriptions, also help optimize prompts, improving output alignment with user intent~\citep{paraphraselm_2024}. However, these approaches typically rely on large models and high computational costs.
In contrast, MPR leverages smaller, task-specific models to refine prompts efficiently, correcting common issues like punctuation, typographical errors, and misuse of key terms. This lightweight approach enhances prompt quality across LLM architectures without requiring significant computational resources.

\section{Multi-stage Prompt Refinement} \label{sec3}

\begin{figure*}[t] \centerline{\includegraphics[width=1\textwidth]{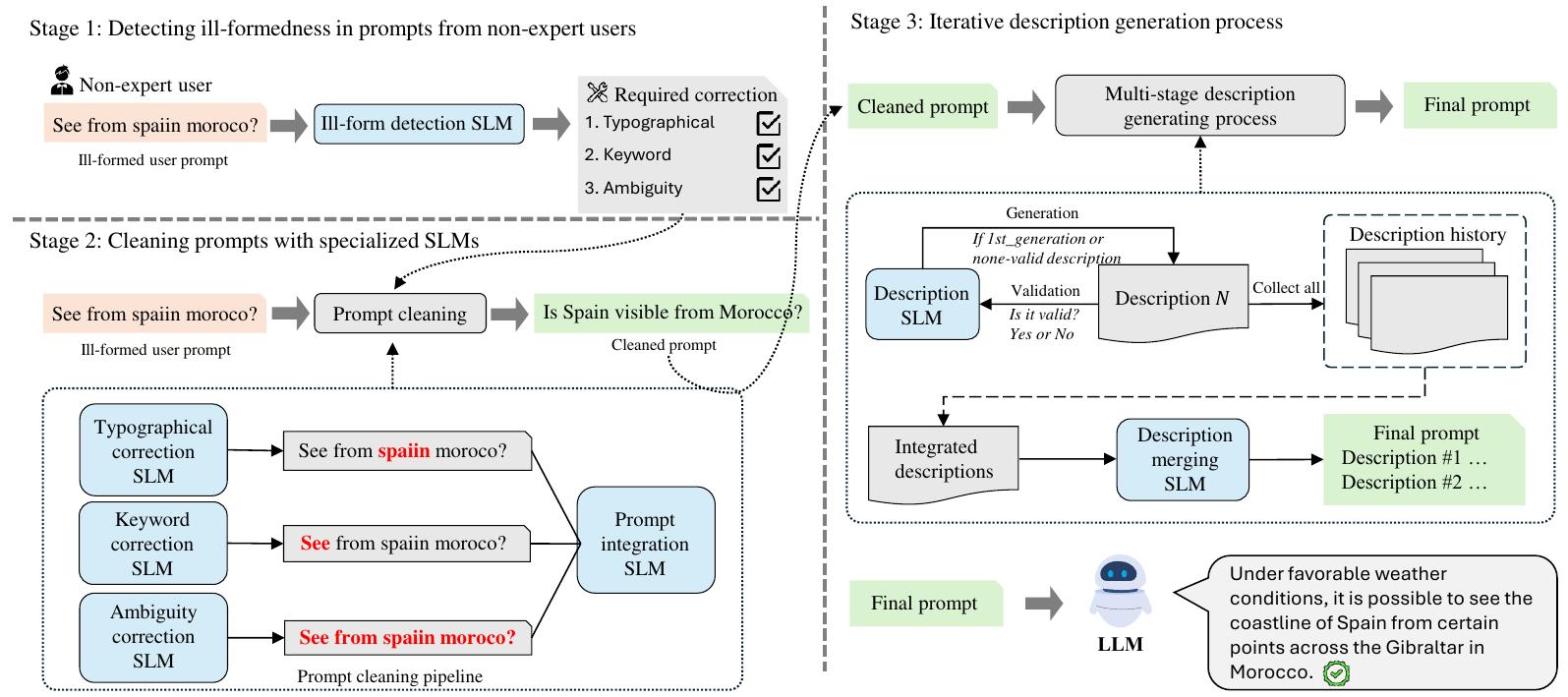}} \caption{Overview of the MPR framework. MPR is designed to systematically refine user prompts before they are processed by LLMs. The framework addresses common errors in prompts such as punctuation mistakes, typographical issues, and misuse of key terms that can lead to inaccurate or suboptimal outputs from LLMs. } \label{fig} \end{figure*}

\subsection{Dataset Construction for Fine-tuning} \label{subsec31}

A critical requirement for training the SLMs in MPR is the construction of a well-curated and diverse dataset that addresses the various aspects of prompt refinement. The dataset is built from four primary sources: the Online Language Modeling (OLM) Wikipedia dataset~\citep{wikipedia_2024}, the Grammarly CoEdIT dataset (CoEdIT)~\citep{CoEdit_2023} ,the Multi-domain Question Rewriting dataset (MQR)~\citep{betterquestions_2020}, and the Magpie-Pro-300K-Filtered dataset (Magpie)~\citep{magpie_2024}. Each dataset serves a unique purpose in training the model for tasks such as punctuation correction, grammar correction, paraphrasing, and description generation.

\noindent{}\textbf{The OLM Wikipedia dataset}, sourced from high-quality English Wikipedia text, is ideal for fine-tuning SLMs on punctuation correction and grammar refinement. Using GPT-3.5-turbo, we curated 10,000 grammatically perfect entries, serving as the gold standard in our training pipeline. To aid learning, we introduced systematic punctuation errors, creating a paired dataset: one version with mistakes and the other as the corrected reference. 

\noindent{}\textbf{The CoEdIT dataset}, drawn from high-quality English text, focuses on non-meaning-changing edits such as fluency, coherence, and style, making it ideal for training models in grammatical and stylistic refinement. Unlike typical datasets, CoEdIT uses instruction-based prompts to guide edits like grammar correction and paraphrasing. We developed 10,000 grammatically accurate entries as the gold standard for well-formed text. To enhance error correction capabilities, we created a parallel dataset with deliberate grammatical mistakes paired with corrected versions.

\noindent{}\textbf{The MQR dataset} contains 2,114 pairs of paraphrased questions across various domains, making it a key resource for training models to handle nuanced linguistic transformations beyond simple synonym swaps. It helps models recognize semantic equivalence between different sentence structures while preserving meaning. For example, “What is the capital of France?” might be paraphrased as “Could you tell me which city serves as the capital of France?”

\noindent{}\textbf{The Magpie Dataset}, consisting of 300,000 keyword-description pairs, trains models to generate concise, accurate descriptions for specific terms, including technical jargon and abbreviations. It helps models interpret complex terminology and produce clear explanations. For instance, when given ``ViT" (Vision Transformer), the model should generate a description like ``ViT is a neural network architecture used for image classification." This improves the model’s ability to clarify ambiguous terms and provide detailed, contextually relevant responses.

\subsection{Training for Prompt Refinement} \label{subsec32}

After constructing the dataset, the SLMs are fine-tuned for specific prompt refinement tasks using instruction-based fine-tuning combined with 4-bit quantized low-rank adaptation (QLoRA)~\citep{qlora_2024}. This method ensures efficient and robust fine-tuning, allowing the SLM to specialize in prompt refinement without overwriting the pre-trained model's general knowledge.
We chose QLoRA for its ability to reduce memory requirements while maintaining high performance as well as its advantages over full fine-tuning in domain adaptation. By applying 4-bit quantization to model weights, QLoRA lowers the memory footprint, enabling fine-tuning on resource-constrained hardware. It also uses low-rank adaptation layers, updating only a small subset of parameters to capture domain-specific nuances while minimizing computational costs. Despite its efficiency, QLoRA performs comparably to traditional fine-tuning, making it ideal for prompt refinement tasks in SLMs where high-quality output and resource efficiency are crucial. This makes QLoRA a scalable, effective solution for enhancing SLMs' performance in complex prompt refinement.

\subsection{Multi-stage Prompt Cleaning and Paraphrasing} \label{subsec33}

MPR operates in a multi-stage manner, with each stage targeting specific types of errors or refinement tasks. This structured approach ensures that even complex ill-formed prompts are systematically refined by the fine-tuned SLMs, improving both the accuracy and relevance of LLM outputs.

\noindent{}\textbf{Identifying the Stage of Error} 

To identify different levels of errors within a given prompt, we fine-tuned an SLM using a specially curated dataset that contains prompts embedded with varying types of mistakes with its corresponding stages of error as its label. This enabled the classification SLM to learn how to identify and classify the stage of error in any given prompt. The resulting fine-tuned model was capable of categorizing unseen inputs according to the severity and nature of errors into three stages, which provided MPR with the knowledge of which stage of correction was needed.

\noindent{}\textbf{Stage 1: Punctuation Correction} 

The first stage corrects basic punctuation and capitalization errors, which are common in casual or hastily written prompts. For instance, a prompt like ``what is the caPital of fRAnce?" would be refined to ``What is the capital of France?" Improving syntactic clarity at this stage helps the LLM interpret inputs more accurately and prepares the prompt for further refinement.

\noindent{}\textbf{Stage 2: Typographical and Syntactical Error Correction} 

The second stage corrects typographical and syntactical errors that can obscure prompt meaning and lead to hallucinations. The SLM identifies and fixes misspelled words or incorrect word usage. For example, ``GAM" would be corrected to ``GAN," improving the prompt’s clarity. By addressing these errors, the system helps the LLM generate more accurate and contextually appropriate responses.

\noindent{}\textbf{Stage 3: Semantic Alignment and Paraphrasing} 

The third stage refines the semantics of the prompt through paraphrasing, clarifying vague or ambiguous inputs. For example, ``Tell me about transformers" could be rephrased as ``Can you explain how Transformer-based neural networks work?" This ensures that the LLM aligns its response with the user’s intent depending on the dialogue context, reducing the risk of hallucinations. Without paraphrasing, the model might provide irrelevant information, such as details about the ``Transformers" movie.

\noindent{}\textbf{Iterative Informative Description Generation} 

After cleaning and paraphrasing, the SLM generates supplementary descriptions to clarify ambiguous terms or provide context. For instance, if the prompt includes ``ViT," the SLM might add, ``ViT, or Vision Transformer, is a deep learning model used for image recognition tasks." This helps the LLM generate more accurate responses. We iteratively generate these by asking the SLM if the information is sufficient and concise enough for generating a satisfactory answer, inducing the model to self-reflection. Descriptions are then ranked by perplexity scores, with the lowest scores indicating the most coherent and relevant content, ensuring that only the most contextually appropriate descriptions are appended to the prompt. Also, if the refined prompt is enough for generation, descriptions are not generated at all, leading to efficiency in time and computational cost.

\noindent{}\textbf{Summary of MPR} 

MPR combines multi-stage prompt cleaning, paraphrasing, and informative description generation into a robust framework for mitigating hallucinations. By leveraging SLMs and task-specific fine-tuning, MPR systematically improves prompt quality across multiple levels. The iterative generation of descriptions ensures that the input to the LLM is not only accurate but also contextually relevant, leading to higher-quality, more reliable outputs across a wide range of applications.

\section{Experiments}
\label{sec4}

We evaluate MPR across five key aspects: 1) the overall performance of refined prompts, 2) MPR’s cleaning and paraphrasing abilities, 3) the quality of generated descriptions, 4) comparisons with existing hallucination mitigation frameworks, and 5) ablation studies exploring the effectiveness of MPR in conjunction with newer hallucination mitigation techniques. This section outlines the details of each experiment, including implementation specifics, evaluation metrics, comprehensive results, and additional analyses conducted on various datasets and models.

\subsection{Experimental Settings}
\label{subsec41}

\subsubsection{Evaluation Dataset}
\label{subsubsec411}

To evaluate MPR's ability to handle ill-formed prompts, we employed a dataset comprising 8,000 user queries from the Google Well-formed Query dataset~\citep{wellformedqueries_2018}. Each query in this dataset had a well-formedness score below 0.5, reflecting suboptimal prompt quality. We ensured that all queries were anonymized and randomized to eliminate biases in processing and evaluation.
Additionally, we extended our evaluation by testing MPR on three widely-used question-answering (QA) datasets: Grade School Math 8K (GSM8K)~\citep{gsm8k_2021}, The Stanford Question Answering Dataset (SQuAD)~\citep{squad_2016}, and Natural Questions (NQ)~\citep{naturalquestions_2019}. 
These datasets were selected for their diversity in query formats and popularity in evaluating LLM performance, allowing us to comprehensively assess MPR's ability to refine prompts across various levels of ill-formedness. To further stress-test the framework, we deliberately introduced errors into the queries in three stages of sabotage:

\begin{table}[t!]
\centering
\renewcommand{\arraystretch}{1}
\setlength{\tabcolsep}{29pt}
\scriptsize
\caption{
Performance of sabotaged queries across stages evaluated with the GPT-3.5-turbo API: Comparison of the Hallucination Index (HI) and Content Quality Score (CQS) for each stage of sabotaged queries before and after applying MPR. All metrics were evaluated using the GPT-3.5-turbo API as a judge, highlighting the effectiveness of the sabotaged prompts.}
\label{table:sabotaged-performance}
\begin{tabular}{cccc}
\toprule
\textbf{Dataset}          & \textbf{Stage}    & \textbf{HI $\downarrow$} & \textbf{CQS $\uparrow$} \\ \midrule
\multirow{4}{*}{Well-formed Queries~\citep{wellformedqueries_2018}} 
                          & Stage 1   & 0.25                             & 0.78                                \\ \cmidrule{2-4}
                          & Stage 2   & 0.35                             & 0.65                                \\ \cmidrule{2-4}
                          & Stage 3   & 0.48                             & 0.55                                \\ \midrule
                          
\multirow{4}{*}{GSM8K~\citep{gsm8k_2021}}    
                          & Stage 1   & 0.27                             & 0.75                                \\ \cmidrule{2-4}
                          & Stage 2   & 0.36                             & 0.64                                \\ \cmidrule{2-4}
                          & Stage 3   & 0.47                             & 0.53                                \\ \midrule
                          
\multirow{4}{*}{SQuAD~\citep{squad_2016}}    
                          & Stage 1    & 0.28                             & 0.74                                \\ \cmidrule{2-4}
                          & Stage 2    & 0.37                             & 0.62                                \\ \cmidrule{2-4}
                          & Stage 3    & 0.49                             & 0.50                                \\ \bottomrule
\end{tabular}
\end{table}

\begin{itemize}
    \item \textbf{Stage 1 Sabotage:} Basic punctuation errors (e.g., missing commas, periods, inconsistent capitalization) were introduced into the prompts.
    \item \textbf{Stage 2 Sabotage:} In addition to punctuation issues, typographical errors (e.g., misspellings, grammatical mistakes) were introduced into the prompts.
    \item \textbf{Stage 3 Sabotage:} We further altered key technical terms, abbreviations, or domain-specific jargon (e.g., changing ``BERT" to ``VERT" or ``GAN " to ``GAM") to evaluate MPR’s ability to detect and correct these complex misinputs.
\end{itemize}

Each dataset was sabotaged across these three stages, and we assessed MPR’s performance in refining these queries at each level of ill-formedness. The effectiveness of the sabotages are illustrated in Table~\ref{table:sabotaged-performance}.

\subsubsection{Hardware Settings}
\label{subsubsec412}

We conducted fine-tuning of the SLMs on a single NVIDIA RTX A6000 GPU, while inference tasks were run using a single NVIDIA TITAN V GPU. These hardware configurations were chosen to balance processing power and efficiency, ensuring scalability for both large models and real-time evaluations. The computational setup enabled us to fine-tune the models quickly while maintaining consistent performance for evaluation tasks.

\begin{table}[t!]
\centering
\renewcommand{\arraystretch}{1}
\setlength{\tabcolsep}{11pt} 
\scriptsize
\caption{MPR's performance on the Well-formed Queries dataset~\citep{wellformedqueries_2018}. Comparison of Hallucination Index (HI), Content Quality Score (CQS), and Win Rate (WR) before and after applying MPR, showing lower hallucinations and higher quality content across all stages of sabotage evaluated with the GPT-3.5-turbo API as a judge.}
\label{table:well-formed-mpr-performance}
\begin{tabular*}{\textwidth}{llllc}
\toprule
\textbf{Models} & \textbf{Stage of Sabotage} & \textbf{HI $\downarrow$} & \textbf{CQS $\uparrow$} & \textbf{WR $\uparrow$} \\ \midrule
Baseline (No Refinement)          & -                          & 0.81                        & 0.52                     & -        \\ \midrule
\multirow{3}{*}{LLaMA-2 (7B)~\citep{llama2_2023}  }             & Stage 1 (Low)              & 0.26 {\color{Green}(-0.55)}          & 0.80 {\color{Green}(+0.28)}           & 91\%        \\  
                            & Stage 2 (Medium)           & 0.37 {\color{Green}(-0.44)}          & 0.69 {\color{Green}(+0.17)}           & 89\%        \\  
                            & Stage 3 (High)             & 0.48 {\color{Green}(-0.33)}          & 0.60 {\color{Green}(+0.08)}           & 86\%        \\ \midrule
                            
\multirow{3}{*}{Phi-3 (3.8B)~\citep{phi3_2024}}               & Stage 1 (Low)              & 0.28 {\color{Green}(-0.53)}          & 0.78 {\color{Green}(+0.26)}           & 90\%        \\  
                            & Stage 2 (Medium)           & 0.36 {\color{Green}(-0.45)}          & 0.67 {\color{Green}(+0.15)}           & 88\%        \\  
                            & Stage 3 (High)             & 0.47 {\color{Green}(-0.34)}          & 0.58 {\color{Green}(+0.06)}           & 84\%        \\ \midrule
                            
\multirow{3}{*}{LLaMA-3.2 (3.21B)~\citep{llama32_2024} }          & Stage 1 (Low)              & 0.25 {\color{Green}(-0.56)}          & 0.82 {\color{Green}(+0.30)}           & 92\%        \\  
                            & Stage 2 (Medium)           & 0.34 {\color{Green}(-0.47)}          & 0.70 {\color{Green}(+0.18)}           & 90\%        \\  
                            & Stage 3 (High)             & 0.45 {\color{Green}(-0.36)}          & 0.62 {\color{Green}(+0.10)}           & 88\%        \\ \midrule
                            
\multirow{3}{*}{Qwen-2.5 (3B)~\citep{qwen2.5_2024}}               & Stage 1 (Low)              & 0.26 {\color{Green}(-0.55)}          & 0.79 {\color{Green}(+0.27)}           & 89\%        \\  
                            & Stage 2 (Medium)           & 0.36 {\color{Green}(-0.45)}          & 0.67 {\color{Green}(+0.15)}           & 87\%        \\  
                            & Stage 3 (High)             & 0.48 {\color{Green}(-0.33)}          & 0.56 {\color{Green}(+0.04)}           & 85\%        \\ \midrule
                        
\multirow{3}{*}{Phi-2 (2.7B)~\citep{phi_2023}}               & Stage 1 (Low)              & 0.29 {\color{Green}(-0.52)}          & 0.77 {\color{Green}(+0.25)}           & 87\%        \\  
                            & Stage 2 (Medium)           & 0.37 {\color{Green}(-0.44)}          & 0.65 {\color{Green}(+0.13)}           & 85\%        \\  
                            & Stage 3 (High)             & 0.49 {\color{Green}(-0.32)}          & 0.54 {\color{Green}(+0.02)}           & 83\%        \\ \midrule
                            
\multirow{3}{*}{Gemma-2 (2B)~\citep{gemma_2024}}                 & Stage 1 (Low)              & 0.28 {\color{Green}(-0.53)}          & 0.76 {\color{Green}(+0.24)}           & 88\%        \\  
                            & Stage 2 (Medium)           & 0.38 {\color{Green}(-0.43)}          & 0.65 {\color{Green}(+0.13)}           & 86\%        \\  
                            & Stage 3 (High)             & 0.50 {\color{Green}(-0.31)}          & 0.54 {\color{Green}(+0.02)}           & 84\%        \\ \midrule\midrule

\textbf{Average}             & -                          & 0.37 {\color{Green}(-0.44)}          & 0.68 {\color{Green}(+0.16)}           & 86\%      \\ \bottomrule
\end{tabular*}
\end{table}

\begin{table}[t!]
\centering
\renewcommand{\arraystretch}{1}
\setlength{\tabcolsep}{11pt} 
\scriptsize
\caption{MPR's performance on the GSM8K dataset~\citep{gsm8k_2021}. Comparison of Hallucination Index (HI), Content Quality Score (CQS), and Win Rate (WR) before and after applying MPR, showing lower hallucinations and higher quality content across all stages of sabotage evaluated with the GPT-3.5-turbo API as a judge.}
\label{table:gsm8k-mpr-performance}
\begin{tabular*}{\textwidth}{llllc}
\toprule
\textbf{Model used for MPR} & \textbf{Stage of Sabotage} & \textbf{HI $\downarrow$} & \textbf{CQS $\uparrow$} & \textbf{WR $\uparrow$} \\ \midrule
Baseline (No Refinement)          & -                          & 0.84                        & 0.47                     & 72\%        \\ \midrule
\multirow{3}{*}{LLaMA-2 (7B)~\citep{llama2_2023}}               & Stage 1 (Low)              & 0.50 {\color{Green}(-0.34)}          & 0.65 {\color{Green}(+0.18)}           & 82\%        \\  
                            & Stage 2 (Medium)           & 0.55 {\color{Green}(-0.29)}          & 0.60 {\color{Green}(+0.13)}           & 80\%        \\  
                            & Stage 3 (High)             & 0.60 {\color{Green}(-0.24)}          & 0.50 {\color{Green}(+0.03)}           & 78\%        \\ \midrule
\multirow{3}{*}{Phi-3 (3.8B)~\citep{phi3_2024}}               & Stage 1 (Low)              & 0.48 {\color{Green}(-0.36)}          & 0.66 {\color{Green}(+0.19)}           & 81\%        \\  
                            & Stage 2 (Medium)           & 0.45 {\color{Green}(-0.39)}          & 0.63 {\color{Green}(+0.16)}           & 79\%        \\  
                            & Stage 3 (High)             & 0.58 {\color{Green}(-0.26)}          & 0.52 {\color{Green}(+0.05)}           & 77\%        \\ \midrule
                            
\multirow{3}{*}{LLaMA-3.2 (3.21B)~\citep{llama32_2024}}          & Stage 1 (Low)              & 0.49 {\color{Green}(-0.35)}          & 0.67 {\color{Green}(+0.20)}           & 83\%        \\  
                            & Stage 2 (Medium)           & 0.50 {\color{Green}(-0.34)}          & 0.63 {\color{Green}(+0.16)}           & 81\%        \\  
                            & Stage 3 (High)             & 0.54 {\color{Green}(-0.30)}          & 0.55 {\color{Green}(+0.08)}           & 79\%        \\ \midrule
\multirow{3}{*}{Qwen-2.5 (3B)~\citep{qwen2.5_2024}}               & Stage 1 (Low)              & 0.51 {\color{Green}(-0.33)}          & 0.64 {\color{Green}(+0.17)}           & 80\%        \\  
                            & Stage 2 (Medium)           & 0.53 {\color{Green}(-0.31)}          & 0.61 {\color{Green}(+0.14)}           & 78\%        \\  
                            & Stage 3 (High)             & 0.59 {\color{Green}(-0.25)}          & 0.52 {\color{Green}(+0.05)}           & 76\%        \\ \midrule

\multirow{3}{*}{Phi-2 (2.7B)~\citep{phi_2023}}               & Stage 1 (Low)              & 0.45 {\color{Green}(-0.39)}          & 0.68 {\color{Green}(+0.21)}           & 79\%        \\  
                            & Stage 2 (Medium)           & 0.52 {\color{Green}(-0.32)}          & 0.60 {\color{Green}(+0.13)}           & 77\%        \\  
                            & Stage 3 (High)             & 0.57 {\color{Green}(-0.27)}          & 0.50 {\color{Green}(+0.03)}           & 75\%        \\ \midrule
                            
\multirow{3}{*}{Gemma-2 (2B)~\citep{gemma_2024}}                 & Stage 1 (Low)              & 0.46 {\color{Green}(-0.38)}          & 0.69 {\color{Green}(+0.22)}           & 78\%        \\  
                            & Stage 2 (Medium)           & 0.54 {\color{Green}(-0.30)}          & 0.60 {\color{Green}(+0.13)}           & 76\%        \\  
                            & Stage 3 (High)             & 0.59 {\color{Green}(-0.25)}          & 0.52 {\color{Green}(+0.05)}           & 75\%        \\ \midrule\midrule

\textbf{Average}             & -                          & 0.52 {\color{Green}(-0.32)}          & 0.61 {\color{Green}(+0.14)}           & 78\%    \\ \bottomrule
\end{tabular*}
\end{table}

\begin{table}[t!]
\centering
\renewcommand{\arraystretch}{1}
\setlength{\tabcolsep}{11pt} 
\scriptsize
\caption{MPR's performance on the SQuAD dataset~\citep{squad_2016}. Comparison of Hallucination Index (HI), Content Quality Score (CQS), and Win Rate (WR) before and after applying MPR, showing lower hallucinations and higher quality content across all stages of sabotage evaluated with the GPT-3.5-turbo API as a judge.}
\label{table:squad-mpr-performance}
\begin{tabular}{llllc}
\toprule
\textbf{Model used for MPR} & \textbf{Stage of Sabotage} & \textbf{HI $\downarrow$} & \textbf{CQS $\uparrow$} & \textbf{WR $\uparrow$} \\ \midrule
Baseline (No Refinement)          & -                          & 0.76                        & 0.57                     & 76\%        \\ \midrule
\multirow{3}{*}{LLaMA-2 (7B)~\citep{llama2_2023}}               & Stage 1 (Low)              & 0.33 {\color{Green}(-0.43)}          & 0.72 {\color{Green}(+0.15)}           & 88\%        \\  
                            & Stage 2 (Medium)           & 0.41 {\color{Green}(-0.35)}          & 0.61 {\color{Green}(+0.04)}           & 86\%        \\  
                            & Stage 3 (High)             & 0.53 {\color{Green}(-0.23)}          & 0.50 {\color{red}(-0.07)}           & 83\%        \\ \midrule
                            
\multirow{3}{*}{Phi-3 (3.8B)~\citep{phi3_2024}}               & Stage 1 (Low)              & 0.34 {\color{Green}(-0.42)}          & 0.70 {\color{Green}(+0.13)}           & 87\%        \\  
                            & Stage 2 (Medium)           & 0.42 {\color{Green}(-0.34)}          & 0.60 {\color{Green}(+0.03)}           & 85\%        \\  
                            & Stage 3 (High)             & 0.54 {\color{Green}(-0.22)}          & 0.49 {\color{red}(-0.08)}           & 82\%        \\ \midrule
                            
\multirow{3}{*}{LLaMA-3.2 (3.21B)~\citep{llama32_2024}}          & Stage 1 (Low)              & 0.32 {\color{Green}(-0.44)}          & 0.74 {\color{Green}(+0.17)}           & 89\%        \\  
                            & Stage 2 (Medium)           & 0.43 {\color{Green}(-0.33)}          & 0.63 {\color{Green}(+0.06)}           & 88\%        \\  
                            & Stage 3 (High)             & 0.52 {\color{Green}(-0.24)}          & 0.50 {\color{red}(-0.07)}           & 85\%        \\ \midrule

\multirow{3}{*}{Qwen-2.5 (3B)~\citep{qwen2.5_2024}}              & Stage 1 (Low)              & 0.34 {\color{Green}(-0.42)}          & 0.73 {\color{Green}(+0.16)}           & 87\%        \\  
                            & Stage 2 (Medium)           & 0.42 {\color{Green}(-0.34)}          & 0.60 {\color{Green}(+0.03)}           & 86\%        \\ 
                            & Stage 3 (High)             & 0.52 {\color{Green}(-0.24)}          & 0.49 {\color{red}(-0.08)}           & 84\%        \\ \midrule
                            
\multirow{3}{*}{Phi-2 (2.7B)~\citep{phi_2023}}              & Stage 1 (Low)              & 0.35 {\color{Green}(-0.41)}          & 0.70 {\color{Green}(+0.13)}           & 79\%        \\ 
                            & Stage 2 (Medium)           & 0.43 {\color{Green}(-0.33)}          & 0.59 {\color{Green}(+0.02)}           & 77\%        \\ 
                            & Stage 3 (High)             & 0.53 {\color{Green}(-0.23)}          & 0.48 {\color{red}(-0.09)}           & 75\%        \\ \midrule
                            
\multirow{3}{*}{Gemma-2 (2B)~\citep{gemma_2024}}               & Stage 1 (Low)              & 0.34 {\color{Green}(-0.42)}          & 0.71 {\color{Green}(+0.14)}           & 78\%        \\  
                            & Stage 2 (Medium)           & 0.42 {\color{Green}(-0.34)}          & 0.60 {\color{Green}(+0.03)}           & 76\%        \\  
                            & Stage 3 (High)             & 0.53 {\color{Green}(-0.23)}          & 0.48 {\color{red}(-0.09)}           & 75\%        \\ \midrule\midrule

\textbf{Average}             & -                          & 0.44 {\color{Green}(-0.32)}          & 0.61 {\color{Green}(+0.04)}           & 81\%      \\ \bottomrule
\end{tabular}
\end{table}

\begin{table}[t!]
\centering
\renewcommand{\arraystretch}{1}
\setlength{\tabcolsep}{11pt} 
\scriptsize
\caption{MPR's performance on the Natural Questions dataset~\citep{naturalquestions_2019}. Comparison of Hallucination Index (HI), Content Quality Score (CQS), and Win Rate (WR) before and after applying MPR, showing lower hallucinations and higher quality content across all stages of sabotage evaluated with the GPT-3.5-turbo API as a judge.}
\label{table:nq-mpr-performance}
\begin{tabular}{llllc}
\toprule
\textbf{Model used for MPR} & \textbf{Stage of Sabotage} & \textbf{HI $\downarrow$} & \textbf{CQS $\uparrow$} & \textbf{WR $\uparrow$} \\ \midrule
Baseline (No Refinement)          & -                          & 0.67                        & 0.62                     & 74\%        \\ \midrule
\multirow{3}{*}{LLaMA-2 (7B)~\citep{llama2_2023}}               & Stage 1 (Low)              & 0.30 {\color{Green}(-0.37)}          & 0.80 {\color{Green}(+0.18)}           & 91\%        \\ 
                            & Stage 2 (Medium)           & 0.35 {\color{Green}(-0.32)}          & 0.70 {\color{Green}(+0.08)}           & 90\%        \\ 
                            & Stage 3 (High)             & 0.45 {\color{Green}(-0.22)}          & 0.60 {\color{red}(-0.02)}           & 87\%        \\ \midrule

\multirow{3}{*}{Phi-3 (3.8B)~\citep{phi3_2024}}               & Stage 1 (Low)              & 0.29 {\color{Green}(-0.38)}          & 0.82 {\color{Green}(+0.20)}           & 92\%        \\ 
                            & Stage 2 (Medium)           & 0.36 {\color{Green}(-0.31)}          & 0.66 {\color{Green}(+0.04)}           & 89\%        \\ 
                            & Stage 3 (High)             & 0.49 {\color{Green}(-0.18)}          & 0.54 {\color{red}(-0.08)}           & 85\%        \\ \midrule
                            
\multirow{3}{*}{LLaMA-3.2 (3.21B)~\citep{llama32_2024}}           & Stage 1 (Low)              & 0.28 {\color{Green}(-0.39)}          & 0.81 {\color{Green}(+0.19)}           & 90\%        \\ 
                            & Stage 2 (Medium)           & 0.38 {\color{Green}(-0.29)}          & 0.64 {\color{Green}(+0.02)}           & 88\%        \\ 
                            & Stage 3 (High)             & 0.50 {\color{Green}(-0.17)}          & 0.56 {\color{red}(-0.06)}           & 86\%        \\ \midrule
                            
\multirow{3}{*}{Qwen-2.5 (3B)~\citep{qwen2.5_2024}}               & Stage 1 (Low)              & 0.30 {\color{Green}(-0.37)}          & 0.75 {\color{Green}(+0.13)}           & 88\%        \\ 
                            & Stage 2 (Medium)           & 0.39 {\color{Green}(-0.28)}          & 0.63 {\color{Green}(+0.01)}           & 87\%        \\ 
                            & Stage 3 (High)             & 0.48 {\color{Green}(-0.19)}          & 0.55 {\color{red}(-0.07)}           & 85\%        \\ \midrule

\multirow{3}{*}{Phi-2 (2.7B)~\citep{phi_2023}}               & Stage 1 (Low)              & 0.31 {\color{Green}(-0.36)}          & 0.74 {\color{Green}(+0.12)}           & 86\%        \\ 
                            & Stage 2 (Medium)           & 0.40 {\color{Green}(-0.27)}          & 0.60 {\color{red}(-0.02)}           & 83\%        \\ 
                            & Stage 3 (High)             & 0.51 {\color{Green}(-0.16)}          & 0.50 {\color{red}(-0.12)}           & 81\%        \\ \midrule
                            
\multirow{3}{*}{Gemma-2 (2B)~\citep{gemma_2024}}                 & Stage 1 (Low)              & 0.32 {\color{Green}(-0.35)}          & 0.73 {\color{Green}(+0.11)}           & 85\%        \\ 
                            & Stage 2 (Medium)           & 0.41 {\color{Green}(-0.26)}          & 0.62 {\color{Green}(0.00)}           & 82\%        \\ 
                            & Stage 3 (High)             & 0.50 {\color{Green}(-0.17)}          & 0.50 {\color{red}(-0.12)}           & 80\%        \\ \midrule\midrule

\textbf{Average}             & -                          & 0.41 {\color{Green}(-0.26)}          & 0.63 {\color{Green}(+0.01)}           & 83\%      \\ \bottomrule
\end{tabular}
\end{table}

\subsection{Evaluating the Effectiveness of MPR Across Various Models}
\label{subsec42}

To measure the effectiveness of MPR in reducing hallucinations and improving content quality, we employed three core metrics: the Hallucination Index (HI), Content Quality Score (CQS), and Win Rate (WR). 
Each metric was evaluated using the GPT-3.5-turbo API as the primary judge to ensure accurate and consistent assessment.

\begin{itemize}
    \item \textbf{HI}: This metric quantifies the factual accuracy of the generated content on a scale from 0 to 1, where 0 represents perfect factual accuracy (no hallucinations), and 1 indicates completely fabricated content. The HI was calculated by having the GPT-3.5-turbo API evaluate the factual correctness of each response.
    
    \item \textbf{CQS}: This metric measures the relevance, coherence, and overall quality of the generated content, with a score ranging from 0 to 1. A score of 1 indicates highly relevant and coherent content. CQS was determined through evaluations by the GPT-3.5-turbo API to ensure consistency and contextual alignment of responses.
    
    \item \textbf{WR}: The win rate measures how often MPR-refined prompts produced better results than the original ill-formed prompts. We compute WR by comparing MPR-refined prompts against original prompts, both with and without description generation, using the GPT-3.5-turbo API to evaluate the effectiveness of MPR in improving the quality of generated responses.
\end{itemize}

To assess MPR’s performance across a variety of language models, we evaluated it on several SLMs, including:

\begin{itemize}
    \item \textbf{LLaMA-2 (7B)}~\citep{llama2_2023}: Developed by Meta, LLaMA-2 is a 7-billion-parameter model that offers robust performance in text generation and contextual understanding, making it suitable for a variety of NLP tasks.
    
    \item \textbf{Qwen-2.5 (3B)}~\citep{qwen2.5_2024}: Developed by Alibaba, Qwen-2.5 is a 3-billion-parameter model known for its advancements in efficiency and accuracy, providing high-quality results in resource-constrained environments.
    
    \item \textbf{Phi-3 (3.8B)}~\citep{phi3_2024}: Developed by Microsoft, Phi-3 is a 3.8-billion-parameter model that balances size and performance, making it suitable for complex text generation and semantic tasks.
    
    \item \textbf{LLaMA-3.2 (3.21B)}~\citep{llama32_2024}: Developed by Meta, LLaMa-3.2 is a compact 3-billion-parameter model optimized for lightweight deployment with robust language generation capabilities.
    
    \item \textbf{Phi-2 (2.7B)}~\citep{phi_2023}: Developed by Microsoft, Phi-2 is a 2.7-billion-parameter model focusing on high-quality language generation with low memory overhead.
    
    \item \textbf{Gemma-2 (2B)}~\citep{gemma_2024}: Developed by DeepMind, Gemma-2 is a 2-billion-parameter model designed for rapid inference and a variety of NLP tasks while maintaining computational efficiency.
\end{itemize}

We evaluated MPR’s performance on these models using the same metrics (HI, CQS, WR). As shown in Table~\ref{table:well-formed-mpr-performance}, Table~\ref{table:gsm8k-mpr-performance}, Table~\ref{table:squad-mpr-performance}, and Table~\ref{table:nq-mpr-performance}, MPR demonstrated substantial improvements across all metrics and datasets. Its impact was particularly notable under high levels of sabotage, where its refinement capabilities were most effective in minimizing hallucinations and preserving response quality. Moreover, MPR achieved strong results across all models, with win rates consistently exceeding 85\%. Larger models, such as Phi-3 and LLaMA-3.2, benefited the most from the additional context provided by MPR’s description generation step, which further enhanced response quality. This indicates that MPR not only boosts the factual accuracy of generated content but also improves its coherence and relevance, ensuring high-quality outputs across a diverse range of SLMs.

\subsection{Evaluation of Prompt Refinement}
\label{subsec44}

To evaluate MPR’s prompt refinement capabilities, we fine-tuned each model and compared the quality of the refined prompts against their original ill-formed versions. We used the following metrics:

\begin{itemize}
\item \textbf{Bilingual Evaluation Understudy (BLEU)}~\citep{bleu_2002}: BLEU evaluates grammatical accuracy and fluency by measuring n-gram overlap between generated and reference texts.

\item \textbf{Recall-Oriented Understudy for Gisting Evaluation (ROUGE)}~\citep{rouge_2022}: ROUGE evaluates text quality based on recall-oriented n-gram overlap. 

\item \textbf{Metric for Evaluation of Translation with Explicit ORdering (METEOR)}~\citep{meteor_2005}: METEOR measures semantic similarity by considering synonyms, stemming, and word matches. Unlike BLEU, it accounts for meaning-level similarities and correlates better with human judgments in paraphrasing and translation tasks.

\end{itemize}

\begin{table}[t!]
\caption{Evaluation of prompt cleaning and paraphrasing on SLMs fine-tuned using our custom dataset. Comparison of BLEU~\citep{bleu_2002}, ROUGE~\citep{rouge_2022}, and METEOR~\citep{meteor_2005} scores before and after fine-tuning, showing improved n-gram overlap, semantic similarity, and contextual relevance in generated descriptions. The evaluation was conducted using the GPT-3.5-turbo API as the primary judge.}
\centering
\setlength{\tabcolsep}{10pt} 
\renewcommand{\arraystretch}{1}
\scriptsize 
\begin{tabular}{llccc}
\toprule
{\textbf{Model}} & {\textbf{Fine-tuning}} & \textbf{BLEU}~\citep{bleu_2002} & \textbf{ROUGE}~\citep{rouge_2022} & \textbf{METEOR}~\citep{meteor_2005} \\
\midrule

\multirow{2}{*}{Llama-2 (7B)~\citep{llama2_2023}}    & Original  & 12.8 & 48.1 & 31.5 \\
& Fine-tuned  & \textbf{23.1} & 56.2 & \textbf{36.5} \\

\midrule
\multirow{2}{*}{Phi-3 (3.8B)~\citep{phi3_2024}}    & Original  & 13.7 & 48.9 & 32.1 \\
& Fine-tuned  & \underline{22.4} & 54.7 & \underline{35.8} \\

\midrule
\multirow{2}{*}{Llama-3.2 (3.21B)~\citep{llama32_2024}}    & Original  & 13.3 & 47.5 & 31.8 \\
& Fine-tuned  & 21.9 & \textbf{56.4} & 35.3 \\

\midrule
\multirow{2}{*}{Qwen-2.5 (3B)~\citep{qwen2.5_2024}}    & Original  & 11.5 & 45.2 & 30.9 \\
& Fine-tuned  & 19.8 & 53.0 & 33.5 \\

\midrule
\multirow{2}{*}{Phi-2 (2.7B)~\citep{phi_2023}}    & Original  & 13.1 & 46.6 & 31.7 \\
& Fine-tuned  & 21.7 & \underline{56.3} & 35.6 \\

\midrule
\multirow{2}{*}{Gemma-2 (2B)~\citep{gemma_2024}}     & Original  & 11.2 & 45.3 & 31.3 \\
& Fine-tuned  & 21.1 & 54.2 & 32.1 \\

\bottomrule
\end{tabular}
\label{table:refinement-quality}
\end{table}
MPR consistently improved the BLEU, ROUGE, and METEOR scores, demonstrating that the framework significantly enhances both linguistic and semantic quality across all sabotage stages and datasets. Improvements in METEOR scores particularly highlight MPR’s ability to align the refined prompts with user intent, as shown in Table~\ref{table:refinement-quality}.

\subsection{Evaluation of Generated Descriptions}
\label{subsec45}
In our extended evaluation of MPR, we focused on assessing two crucial aspects of the descriptions generated by the system: relevance and coherence. 
The evaluation was carried out using the same set of SLMs that had been fine-tuned for prompt refinement. By analyzing these metrics, we aimed to better understand how well MPR improved the quality of the refined prompts, ensuring that they are not only factually correct but also contextually appropriate and logically structured. For the evaluation metrics, we used the following:

\begin{itemize}
    \item Relevance: This metric measures how closely the generated descriptions align with the refined prompt and fulfill the user's intent. It ensures the descriptions provide accurate and meaningful information relevant to the prompt. We used the GPT-3.5-turbo API to rate each description's appropriateness and relevance to the prompt.
    \item Coherence: Coherence assesses the logical structure and flow of the descriptions in relation to the refined prompt. It checks for internal consistency and how seamlessly the descriptions integrate with the prompt. Like relevance, coherence was evaluated using the GPT-3.5-turbo API.
\end{itemize}

Both relevance and coherence were assessed systematically using the GPT-3.5-turbo API. This allowed for an unbiased, automated evaluation process, where the API provided ratings based on pre-defined criteria related to prompt quality. Each generated description was reviewed in the context of the refined prompt, ensuring that it not only aligned with the user's intent (relevance) but also maintained a logical, smooth flow that improved the clarity and usability of the refined text (coherence).

\begin{table}[t!]
\centering
\renewcommand{\arraystretch}{0.8}  
\setlength{\tabcolsep}{9pt}       
\scriptsize
\caption{Evaluation of generated descriptions across models and datasets. This table compares relevance and coherence scores for descriptions generated by various models across multiple datasets. Each description was assessed using the GPT-3.5-turbo API. }
\label{table:generation-quality}
\begin{tabular}{llcc}
\toprule
\textbf{Model}       & \textbf{Dataset}             & \textbf{Relevance Score} & \textbf{Coherence Score} \\ \midrule

\multirow{5}{*}{LLaMA-2 (7B)~\citep{llama2_2023}}              & Well-formed Queries~\citep{wellformedqueries_2018}          & \textbf{0.90}                     & \underline{0.88}                    \\ 
                     & GSM8K~\citep{gsm8k_2021}                        & \underline{0.89}                     & 0.87                    \\ 
                     & SQuAD~\citep{squad_2016}                        & 0.88                     & 0.85                    \\ 
                     & Natural Questions~\citep{naturalquestions_2019}            & 0.87                     & 0.84                    \\ \midrule

\multirow{5}{*}{Phi-3 (3.8B)~\citep{phi3_2024}}         & Well-formed Queries~\citep{wellformedqueries_2018}          & 0.88                     & 0.86                    \\ 
                     & GSM8K~\citep{gsm8k_2021}                        & 0.87                     & 0.85                    \\ 
                     & SQuAD~\citep{squad_2016}                        & 0.86                     & 0.84                    \\ 
                     & Natural Questions~\citep{naturalquestions_2019}            & 0.85                     & 0.83                    \\ \midrule                     
\multirow{5}{*}{LLaMA-3.2 (3.21B)~\citep{llama32_2024}}     & Well-formed Queries~\citep{wellformedqueries_2018}          & 0.88                     & \textbf{0.89}                    \\  
                     & GSM8K~\citep{gsm8k_2021}                        & 0.87                     & 0.86                    \\  
                     & SQuAD~\citep{squad_2016}                        & 0.86                     & 0.84                    \\  
                     & Natural Questions~\citep{naturalquestions_2019}            & 0.85                     & 0.83                    \\ \midrule

\multirow{5}{*}{Qwen-2.5 (3B)~\citep{qwen2.5_2024}}             & Well-formed Queries~\citep{wellformedqueries_2018}          & 0.85                     & 0.83                    \\  
                     & GSM8K~\citep{gsm8k_2021}                        & 0.84                     & 0.82                    \\  
                     & SQuAD~\citep{squad_2016}                        & 0.83                     & 0.81                    \\  
                     & Natural Questions~\citep{naturalquestions_2019}            & 0.82                     & 0.80                    \\ \midrule

\multirow{5}{*}{Phi-2 (2.7B)~\citep{phi_2023}}                & Well-formed Queries~\citep{wellformedqueries_2018}          & 0.84                     & 0.82                    \\  
                     & GSM8K~\citep{gsm8k_2021}                        & 0.83                     & 0.81                    \\  
                     & SQuAD~\citep{squad_2016}                        & 0.82                     & 0.80                    \\  
                     & Natural Questions~\citep{naturalquestions_2019}            & 0.81                     & 0.79                    \\ \midrule

 \multirow{5}{*}{Gemma-2 (2B)~\citep{gemma_2024}}           & Well-formed Queries~\citep{wellformedqueries_2018}          & 0.83                     & 0.81                    \\  
                     & GSM8K~\citep{gsm8k_2021}                        & 0.82                     & 0.80                    \\  
                     & SQuAD~\citep{squad_2016}                        & 0.81                     & 0.79                    \\  
                     & Natural Questions~\citep{naturalquestions_2019}            & 0.80                     & 0.78                    \\ \bottomrule
\end{tabular}
\end{table}
Through this evaluation, we found that MPR was highly effective in generating descriptions that were both relevant and coherent. The refined prompts were consistently enhanced by descriptions that closely matched the user’s intent and adhered to the logical structure required for readability. This resulted in prompts that were not only factually accurate and aligned with the task at hand but also easy to follow and understand.

As presented in Table~\ref{table:generation-quality}, MPR-generated descriptions scored consistently high across all datasets. The iterative generation and ranking process played a crucial role in enhancing both the relevance and coherence of the final prompt input to the LLM.

\subsection{Comparison of MPR with Other Hallucination Mitigation Frameworks}
\label{subsec46}

In our evaluation of MPR, we compared its performance against post-hoc hallucination mitigation methods like SelfCheckGPT~\citep{selfcheckgpt_2023}, CoVE~\citep{cove_2023}, DRESS~\citep{dress_2024}, and MixAlign~\citep{mixalign_2023}. Using the same metrics (HI, CQS, WR) plus Processing Time, we quantitatively assessed each method’s ability to reduce hallucinations and improve LLM performance.

\begin{table}[t]
\centering
\renewcommand{\arraystretch}{1}  
\setlength{\tabcolsep}{9pt}      
\scriptsize
\caption{Comparison of hallucination mitigation frameworks using key evaluation metrics: Hallucination Index (HI), Content Quality Score (CQS), Win Rate (WR), and Processing Time (T). The evaluation was conducted on sabotaged prompts from the SQuAD dataset~\citep{squad_2016}, with MPR implemented using the LLaMA-3.2 model~\citep{llama32_2024}. All metrics were assessed using the GPT-3.5-turbo API.}
\label{table:baseline}
\begin{tabular}{lcccc}
\toprule
\textbf{Framework}         & \textbf{HI $\downarrow$} & \textbf{CQS $\uparrow$} & \textbf{WR $\uparrow$} & \textbf{T (ms)} \\ \midrule
MPR (Ours)                       & 0.18                              & 0.81                                 & 91\%                    & \textbf{1215}                          \\ \midrule
SelfCheckGPT~\citep{selfcheckgpt_2023}               & 0.22                              & 0.76                                 & 85\%                    & 1541                          \\ 
SelfCheckGPT~\citep{selfcheckgpt_2023} + MPR (Ours)           & \textbf{0.14}                     & \textbf{0.85}                        & \textbf{94\%}           & 1478                         \\ \midrule
CoVE~\citep{cove_2023}                    & 0.21                              & 0.77                                 & 86\%                    & 1491                          \\ 
CoVE~\citep{cove_2023} + MPR (Ours)            & 0.16                     & 0.83                        & 92\%           & 1516                         \\ \midrule
DRESS~\citep{dress_2024}                  & 0.20                              & 0.78                                 & 87\%                    & \underline{1382}                          \\ 
DRESS~\citep{dress_2024} + MPR (Ours)           & \underline{0.15}                     & \underline{0.84}                        & \underline{93\%}           & 1562                          \\ \midrule
MixAlign~\citep{mixalign_2023}       & 0.19                              & 0.79                                 & 88\%                    & 1639                         \\ 
MixAlign~\citep{mixalign_2023} + MPR (Ours)           & 0.17                     & 0.82                        & 90\%           & 1759                          \\ \bottomrule

\end{tabular}
\end{table}
\begin{figure*}[t] \centerline{\includegraphics[width=1\textwidth]{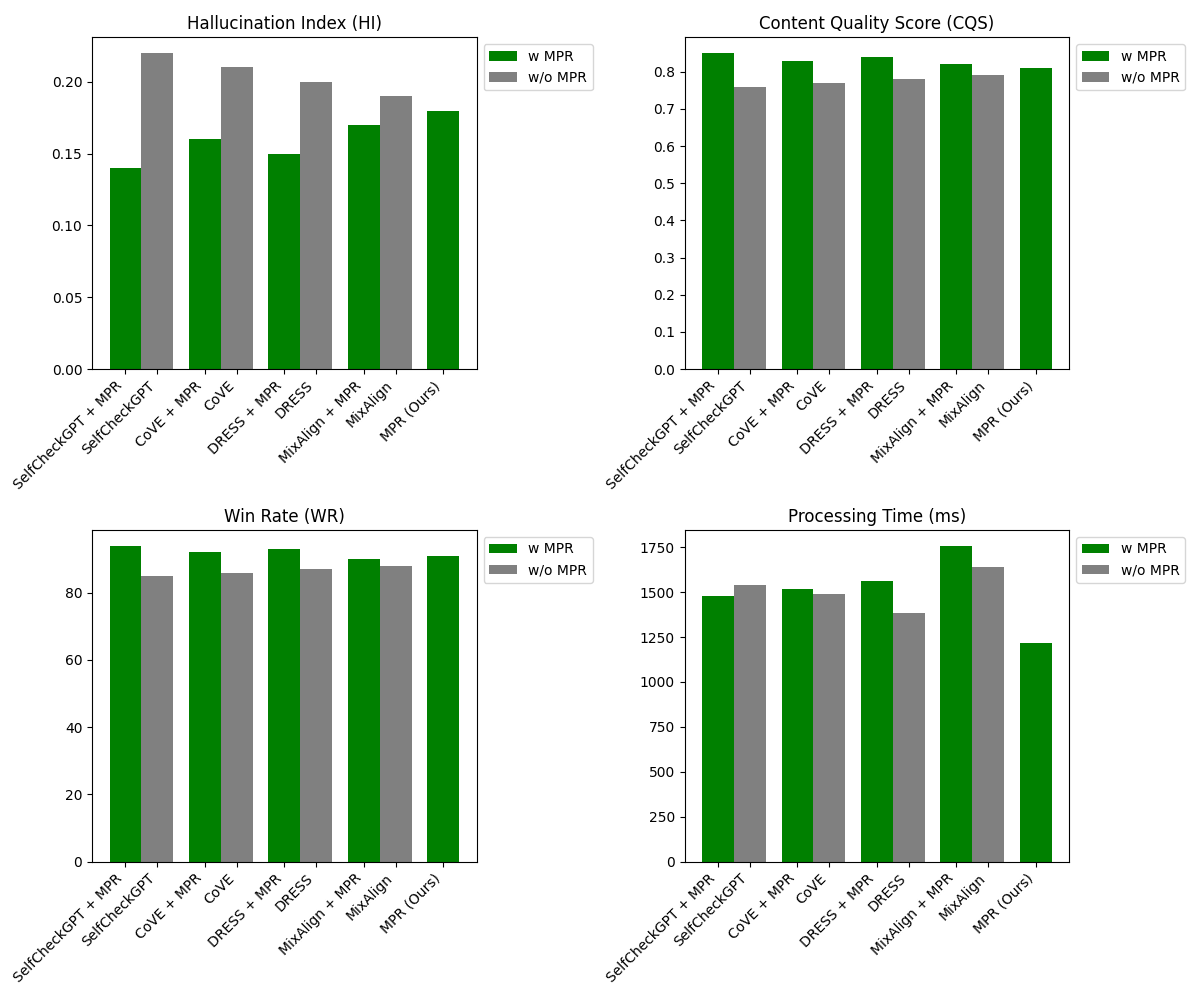}} \caption{Visualization of MPR's effectiveness} \label{fig_2} \end{figure*}

\subsection{Multi-stage Prompt Cleaning and Paraphrasing} \label{subsec48} As shown in Table~\ref{table:baseline} and Figure~\ref{fig_2}, MPR outperformed several baseline methods, particularly in refining prompts and reducing hallucinations. However, in some cases, DRESS and MixAlign delivered comparable or slightly better results. Despite this, MPR remained highly competitive across most evaluations.

One of MPR's key strengths is its model-agnostic design, enabling easy integration with other hallucination mitigation techniques. Combining MPR with these methods led to substantial performance improvements over using any single approach alone. This synergy highlighted MPR’s flexibility and adaptability.

Key takeaways: \begin{itemize} \item MPR is highly effective on its own for reducing hallucinations and improving LLM outputs. \item Combining MPR with other techniques yields even better results, leveraging their complementary strengths for more robust hallucination mitigation. \end{itemize}

In conclusion, MPR is a powerful tool in hallucination mitigation, both as a standalone solution and when used in combination with other state-of-the-art methods. Its adaptability and scalability make it a valuable asset for enhancing the reliability of LLM outputs across diverse applications.

\subsection{Ablation Studies}
\label{subsec47}

\begin{table}[t]
\centering
\renewcommand{\arraystretch}{1}  
\setlength{\tabcolsep}{17pt}       
\scriptsize
\caption{Ablation studies on different configurations of MPR using a single fine-tuned model(LLaMa-3.2 (3B)~\citep{llama32_2024}). Comparison of the Hallucination Index (HI), Content Quality Score (CQS), and Win Rate (WR) across varying MPR configurations. Each configuration was evaluated using the GPT-3.5-turbo API, demonstrating the impact of each setup on reducing hallucinations and enhancing content quality.}
\label{table:ablation}
\begin{tabular}{lccc}
\toprule
\textbf{Configuration}                & \textbf{HI $\downarrow$} & \textbf{CQS $\uparrow$} & \textbf{WR $\uparrow$} \\ \midrule
\textbf{Baseline (No Refinement)}     & 0.30                                        & 0.65                                          & -                  \\ \midrule
Full MPR                              & \textbf{0.14}                                        & \textbf{0.83}                                          & \textbf{93\%}                   \\ 
\text{ }\text{ }w/o Description Generation        & \underline{0.20}                                        & \underline{0.78}                                          & \underline{89\%}                   \\ 
\text{ }\text{ }w/o Multi-stage Cleaning          & 0.24                                        & 0.74                                          & 86\%                   \\ 
\text{ }\text{ }w/o Iterative Ranking           & 0.21                                        & 0.75                                          & 87\%                   \\ \bottomrule
\end{tabular}
\end{table}
\begin{figure*}[t] \centerline{\includegraphics[width=1\textwidth]{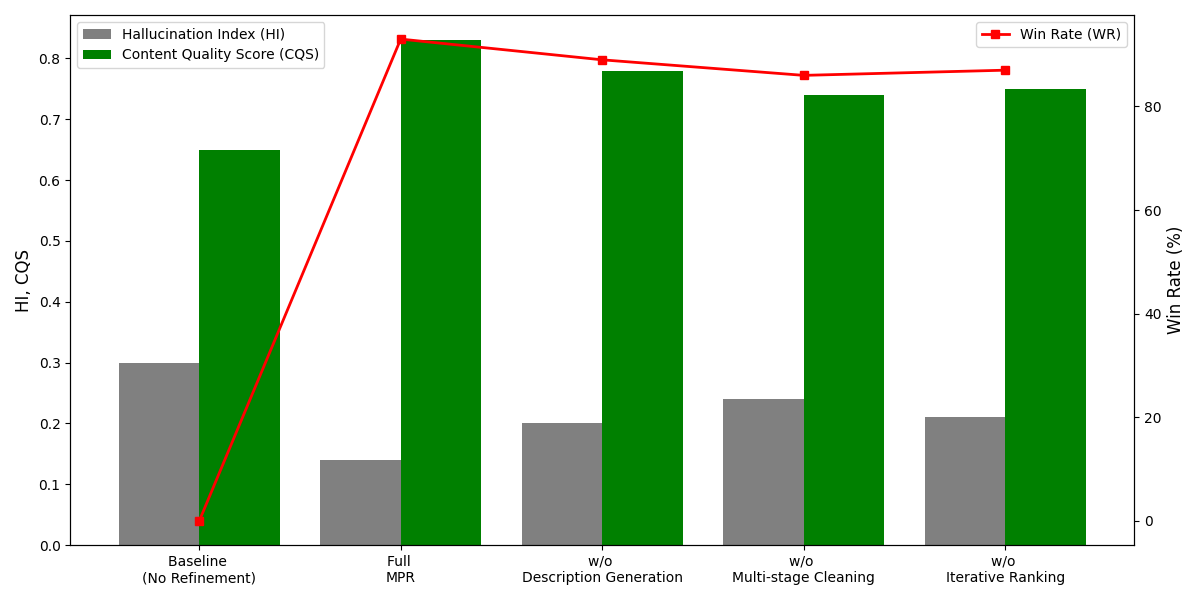}} \caption{Visualization of the importance of each module of MPR. } \label{fig_3} \end{figure*}

In our ablation studies, we conducted experiments to evaluate the impact of MPR’s key components, such as multi-stage prompt cleaning and iterative description generation and ranking. By testing the system with and without these features, we isolated the contribution of each to MPR’s overall performance.

As shown in Table~\ref{table:ablation} and Figure~\ref{fig_3}, these studies confirm MPR’s robustness across various models, datasets, and sabotage stages. The results show that MPR consistently improves refined prompts, leading to several benefits: \begin{itemize} \item Hallucination reduction: MPR effectively minimizes factually incorrect or irrelevant content generated by LLMs. \item Improved output quality: By refining prompts, MPR enhances the accuracy, coherence, and alignment of LLM-generated responses. \end{itemize}

These findings demonstrate MPR’s versatility and scalability, proving its effectiveness across different conditions. Each component contributes meaningfully, validating MPR as a reliable solution for prompt refinement and hallucination mitigation in LLMs.

\section{Limitations}
\label{sec5}

Despite MPR's advancements, several limitations remain that warrant attention.
MPR, trained on general-purpose datasets, may under perform in domain-specific contexts like legal or medical sectors due to specialized jargon. Future work should focus on developing domain-specific fine-tuning strategies to enhance its effectiveness in specialized fields.
While fully automated, MPR could benefit from human-in-the-loop systems to improve the quality of refined prompts. Incorporating human feedback would enhance its ability to capture user intent and address complex nuances, though this would increase system complexity.
Also, current metrics (HI, CQS, WR) primarily evaluate content accuracy and relevance but do not fully capture user satisfaction or fluency. Future research should develop more user-centered evaluation methods to better assess MPR's impact in real-world applications.

\section{Conclusion}
\label{sec61}
In this work, we introduced MPR, a framework designed to mitigate hallucinations in LLMs by refining ill-formed prompts. MPR employs a multi-stage approach to improve clarity, coherence, and task specificity through cleaning, paraphrasing, and generating informative descriptions. Using fine-tuned SLMs, MPR efficiently refines prompts at various levels, from basic error correction to resolving complex semantic issues. Our evaluations across QA datasets (GSM8K, SQuAD, Natural Questions) demonstrate MPR's effectiveness in enhancing LLM output quality and reducing hallucinations.
MPR's modular and scalable design allows it to adapt across different models and domains. However, future work is needed to extend its applicability to specialized fields like law or medicine and multi-modal environments. Incorporating human feedback along with improved evaluation metrics could further improve its performance in handling ambiguous or complex prompts.
In conclusion, MPR significantly advances prompt refinement and hallucination mitigation, offering a scalable solution that enhances LLM performance across diverse applications.
\section*{Acknowledgements}

 This research was supported by the IITP(Institute of Information \& Coummunications Technology Planning \& Evaluation)-ITRC(Information Technology Research Center)(IITP-2024-RS-2024-00436857, 50) grant funded by the Korea government(Ministry of Science and ICT) and the Institute of Information \& communications Technology Planning \& Evaluation (IITP) grant funded by the Korea government(MSIT) (No. 2019-0-00079, Artificial Intelligence Graduate School Program(Korea University)

~\newpage
\bibliographystyle{elsarticle-num}
\bibliography{REFERENCE_Shortened_Journals}

\begin{thebibliography}{10}
\expandafter\ifx\csname url\endcsname\relax
  \def\url#1{\texttt{#1}}\fi
\expandafter\ifx\csname urlprefix\endcsname\relax\def\urlprefix{URL }\fi
\expandafter\ifx\csname href\endcsname\relax
  \def\href#1#2{#2} \def\path#1{#1}\fi

\bibitem{pr_8}
H.~Cheng, Y.~Hehui, X.~Zhou, X.~Liu, F.~Chen, M.~Wang, {Vision-language pre-training via modal interaction}, Pattern Recognit. 156 (2024) 110809.

\bibitem{pr_9}
Z.~Hu, P.~Yang, Y.~Jiang, Z.~Bai., {Prompting large language model with context and pre-answer for knowledge-based VQA}, Pattern Recognit. 151 (2024) 110399.

\bibitem{gpt4_2023}
OpenAI, et~al., {GPT-4 technical report}, arXiv preprint arXiv:2303.08774 (2023).

\bibitem{palm2_2023}
R.~Anil, et~al., {Palm 2 technical report}, arXiv preprint arXiv:2305.10403 (2023).

\bibitem{pr_5}
H.~Wang, G.~Lin, S.~C. Hoi, C.~Miao, Decomposing generation networks with structure prediction for recipe generation, Pattern Recognit. 126 (2022) 108578.

\bibitem{llmsurvey_2023}
W.~X. Zhao, et~al., {A Survey of Large Language Models}, arXiv preprint arXiv:2303.18223 (2023).

\bibitem{llmlies_2023}
J.~Yao, K.~Ning, Z.~Liu, M.~Ning, L.~Yuan, {Llm lies: Hallucinations are not bugs, but features as adversarial examples}, arXiv preprint arXiv:2310.01469 (2023).

\bibitem{pr_3}
F.~Zheng, J.~Cao, W.~Yu, Z.~Chen, N.~Xiao, Y.~Lu, Exploring low-resource medical image classification with weakly supervised prompt learning, Pattern Recognit. 149 (2024) 110250.

\bibitem{optimizers_2024}
R.~Ma, X.~Wang, X.~Zhou, J.~Li, N.~Du, T.~Gui, Q.~Zhang, X.~Huang, {Are Large Language Models Good Prompt Optimizers?}, arXiv preprint arXiv:2402.02101 (2024).

\bibitem{factcheckgpt_2023}
Y.~Wang, et~al., {Factcheck-GPT: End-to-End Fine-Grained Document-Level Fact-Checking and Correction of LLM Output}, arXiv preprint arXiv:2311.09000 (2023).

\bibitem{hallucinationorigin_2023}
N.~Dziri, S.~Milton, M.~Yu, O.~Zaiane, S.~Reddy, {On the Origin of Hallucinations in Conversational Models: Is it the Datasets or the Models?}, in: Proceedings of the North American Chapter of the Association for Computational Linguistics (NAACL), 2023, pp. 5271--5285.

\bibitem{pr_2}
Z.~Cai, H.~Zhang, P.~Zhan, X.~Jia, Y.~Yan, X.~Song, B.~Xie, Multi-schema prompting powered token-feature woven attention network for short text classification, Pattern Recognit. 156 (2024) 110782.

\bibitem{preventinghallucinations_2024}
A.~Gunjal, J.~Yin, E.~Bas, {Detecting and Preventing Hallucinations in Large Vision Language Models}, in: Proceedings of the AAAI Conference on Artificial Intelligence (AAAI), Vol.~38, 2024, pp. 18135--18143.

\bibitem{finetuning_2023}
Z.~Han, C.~Gao, J.~Liu, J.~Zhang, S.~Q. Zhang, {Parameter-Efficient Fine-Tuning for Large Models: A Comprehensive Survey}, Nat. Mach. Intell. 5 (2023) 220–235.

\bibitem{distilling_2023}
C.-Y. Hsieh, et~al., {Distilling Step-by-Step! Outperforming Larger Language Models with Less Training Data and Smaller Model Sizes}, in: Proceedings of the Association for Computational Linguistics (ACL), 2023, pp. 8003--8017.

\bibitem{slmresource_2024}
G.~Bai, et~al., {Beyond Efficiency: A Systematic Survey of Resource-Efficient Large Language Models}, arXiv preprint arXiv:2401.00625 (2024).

\bibitem{differently_2023}
J.~Wei, et~al., {Larger language models do in-context learning differently}, arXiv preprint arXiv:2303.03846 (2023).

\bibitem{pr_6}
D.~Luo, Y.~Liu, R.~Yang, X.~Liu, J.~Zeng, Y.~Zhou, X.~Bai, Toward real text manipulation detection: New dataset and new solution, Pattern Recognit. 157 (2025) 110828.

\bibitem{knowledgeintensive_2024}
M.~Kang, S.~Lee, J.~Baek, K.~Kawaguchi, S.~J. Hwang, {Knowledge-augmented reasoning distillation for small language models in knowledge-intensive tasks}, in: Advances in Neural Information Processing Systems (NeurIPS), Vol.~36, 2024, pp. 48573--48602.

\bibitem{pr_1}
Y.-E. Kim, Y.-W. Lee, S.-W. Lee, {LC-MSM: Language-Conditioned Masked Segmentation Model for unsupervised domain adaptation}, Pattern Recognit. 148 (2024) 110201.

\bibitem{teachingSLMs_2022}
L.~C. Magister, J.~Mallinson, J.~Adamek, E.~Malmi, A.~Severyn, {Teaching small language models to reason}, in: Proceedings of the Association for Computational Linguistics (ACL), 2023, pp. 1773--1781.

\bibitem{pr_7}
X.~Ke, H.~Liu, P.~Xu, X.~Lin, W.~Guo, Text-based person search via cross-modal alignment learning, Pattern Recognit. Lett. 152 (2024) 110481.

\bibitem{SLMsaregood_2024}
P.~Lepagnol, T.~Gerald, S.~Ghannay, C.~Servan, S.~Rosset, {Small Language Models are Good Too: An Empirical Study of Zero-Shot Classification}, arXiv preprint arXiv:2404.11122 (2024).

\bibitem{promptrefinement_2023}
A.~Madaan, et~al., {SELF-REFINE: Iterative Refinement with Self-Feedback}, in: Advances in Neural Information Processing Systems (NeurIPS), Vol.~36, 2023, pp. 46534--46594.

\bibitem{promptrewritingrl_2024}
W.~Kong, S.~A. Hombaiah, M.~Zhang, Q.~Mei, M.~Bendersky, {PRewrite: Prompt Rewriting with Reinforcement Learning}, arXiv preprint arXiv:2401.08189 (2024).

\bibitem{humanfeedback_2024}
S.~Chaudhari, et~al., {RLHF Deciphered: A Critical Analysis of Reinforcement Learning from Human Feedback for LLMs}, arXiv preprint arXiv:2404.08555 (2024).

\bibitem{pr_4}
B.~Han, X.~Jiang, Z.~Fang, H.~Fujita, Y.~Gao, F-scp: An automatic prompt generation method for specific classes based on visual language pre-training models, Pattern Recognit. 147 (2024) 110096.

\bibitem{paraphraselm_2024}
L.~Jayawardena, P.~Yapa, {ParaFusion: A Large-Scale LLM-Driven English Paraphrase Dataset Infused with High-Quality Lexical and Syntactic Diversity}, arXiv preprint arXiv:2404.12010 (2024).

\bibitem{wikipedia_2024}
Y.~Shao, Y.~Jiang, T.~A. Kanell, P.~Xu, O.~Khattab, M.~S. Lam, {Assisting in writing wikipedia-like articles from scratch with large language models}, arXiv preprint arXiv:2402.14207 (2024).

\bibitem{CoEdit_2023}
P.~Jain, et~al., {A Voice Enabled Grammar Correction System using Transformers}, in: International Conference on Artificial Intelligence For Internet of Things (AIIoT), 2023, pp. 1--6.

\bibitem{betterquestions_2020}
Z.~Chu, M.~Chen, J.~Chen, M.~Wang, K.~Gimpel, M.~Faruqui, X.~Si, {How to Ask Better Questions? A Large-Scale Multi-Domain Dataset for Rewriting Ill-Formed Questions}, in: Proceedings of the AAAI Conference on Artificial Intelligence (AAAI), Vol.~34, 2020, pp. 7586--7593.

\bibitem{magpie_2024}
Z.~Xu, F.~Jiang, L.~Niu, Y.~Deng, R.~Poovendran, Y.~Choi, B.~Y. Lin, {Magpie: Alignment Data Synthesis from Scratch by Prompting Aligned LLMs with Nothing}, arXiv preprint arXiv:2406.08464 (2024).

\bibitem{qlora_2024}
T.~Dettmers, A.~Pagnoni, A.~Holtzman, L.~Zettlemoyer, {Qlora: Efficient finetuning of quantized llms}, in: Advances in Neural Information Processing Systems (NeurIPS), Vol.~36, 2024, pp. 10088--10115.

\bibitem{wellformedqueries_2018}
M.~Faruqui, D.~Das, {Identifying Well-formed Natural Language Questions}, in: Proceedings of the Conference on Empirical Methods in Natural Language Processing (EMNLP), 2018, pp. 798--803.

\bibitem{gsm8k_2021}
K.~Cobbe, et~al., {Training verifiers to solve math word problems}, arXiv preprint arXiv:2110.14168 (2021).

\bibitem{squad_2016}
P.~Rajpurkar, et~al., {SQuAD: 100,000+ Questions for Machine Comprehension of Text}, arXiv preprint arXiv:1606.05250 (2016).

\bibitem{naturalquestions_2019}
T.~K. others, Natural questions: a benchmark for question answering research, Trans. Assoc. Comput. Linguist. 7 (2019) 453--466.

\bibitem{llama2_2023}
H.~Touvron, et~al., {Llama 2: Open Foundation and Fine-Tuned Chat Models}, arXiv preprint arXiv:2307.09288 (2023).

\bibitem{phi3_2024}
M.~Abdin, et~al., {Phi-3 technical report: A highly capable language model locally on your phone}, arXiv preprint arXiv:2404.14219 (2024).

\bibitem{llama32_2024}
A.~Dubey, et~al., {The llama 3 herd of models}, arXiv preprint arXiv:2407.21783 (2024).

\bibitem{qwen2.5_2024}
A.~Yang, et~al., {Qwen2 technical report}, arXiv preprint arXiv:2407.10671 (2024).

\bibitem{phi_2023}
Y.~Li, et~al., {Textbooks Are All You Need II: phi-1.5 technical report}, arXiv preprint arXiv:2309.05463 (2023).

\bibitem{gemma_2024}
G.~Team, et~al., {Gemma: Open Models Based on Gemini Research and Technology}, arXiv preprint arXiv:2403.08295 (2024).

\bibitem{bleu_2002}
K.~Papineni, S.~Roukos, T.~Ward, W.-J. Zhu, {BLEU: a Method for Automatic Evaluation of Machine Translation}, in: Proceedings of the Association for Computational Linguistics (ACL), 2002, pp. 311--318.

\bibitem{rouge_2022}
P.~Lu, et~al., {Learn to Explain: Multimodal Reasoning via Thought Chains for Science Question Answering}, in: Advances in Neural Information Processing Systems (NeurIPS), Vol.~35, 2022, pp. 2507--2521.

\bibitem{meteor_2005}
S.~Banerjee, A.~Lavie, {METEOR: An Automatic Metric for MT Evaluation with Improved Correlation with Human Judgments}, in: Proceedings of the Association for Computational Linguistics (ACL), 2005, pp. 65--72.

\bibitem{selfcheckgpt_2023}
P.~Manakul, A.~Liusie, M.~J.~F. Gales, {SELFCHECKGPT: Zero-Resource Black-Box Hallucination Detection for Generative Large Language Models}, in: Proceedings of the Conference on Empirical Methods in Natural Language Processing (EMNLP), 2023, pp. 9004--9017.

\bibitem{cove_2023}
S.~Dhuliawala, M.~Komeili, J.~Xu, R.~Raileanu, X.~Li, A.~Celikyilmaz, J.~Weston, {Chain-of-Verification Reduces Hallucination in Large Language Models}, arXiv preprint arXiv:2309.11495 (2023).

\bibitem{dress_2024}
Y.~Chen, K.~Sikka, M.~Cogswell, H.~Ji, A.~Divakaran, {Dress: Instructing large vision-language models to align and interact with humans via natural language feedback}, in: Proceedings of the IEEE Conference on Computer Vision and Pattern Recognition (CVPR), 2024, pp. 14239--14250.

\bibitem{mixalign_2023}
S.~Zhang, L.~Pan, J.~Zhao, W.~Y. Wang, {The Knowledge Alignment Problem: Bridging Human and External Knowledge for Large Language Models}, arXiv preprint arXiv:2305.13669 (2023).

\end{thebibliography}

\end{document}